\documentclass{article}
\usepackage{ijcai24}
\usepackage{times}
\usepackage{soul}
\usepackage{url}
\usepackage[hidelinks]{hyperref}
\usepackage[utf8]{inputenc}
\usepackage[small]{caption}
\usepackage{graphicx}
\usepackage{amsmath}
\usepackage{amsthm}
\usepackage{booktabs}
\usepackage{algorithm}
\usepackage{algorithmic}
\usepackage[switch]{lineno}
\usepackage{array}
\usepackage{tabularx} % Add this to your preamble for dynamic column widths
\usepackage{booktabs} % Optional: For cleaner table lines (optional but recommended)
\usepackage{enumitem}
\usepackage{float}
\usepackage{listings}
\usepackage{xcolor}
\usepackage{tikz}
\usepackage{amsmath}
\usepackage{algorithm}
\usepackage{algorithmic}
\usepackage{caption}
\usetikzlibrary{arrows.meta, positioning, calc, fit}
\usepackage{stfloats}
\usepackage{subcaption}

\usepackage{listofitems} % for \readlist to create arrays
\usepackage[outline]{contour} % glow around text
\contourlength{1.4pt}

\usepackage{xcolor}
\colorlet{myred}{red!80!black}
\colorlet{myblue}{blue!80!black}
\colorlet{mygreen}{green!60!black}
\colorlet{myorange}{orange!70!red!60!black}
\colorlet{mydarkred}{red!30!black}
\colorlet{mydarkblue}{blue!40!black}
\colorlet{mydarkgreen}{green!30!black}

\tikzset{
  >=latex, % for default LaTeX arrow head
  node/.style={thick,circle,draw=myblue,minimum size=22,inner sep=0.5,outer sep=0.6},
  node in/.style={node,green!20!black,draw=mygreen!30!black,fill=mygreen!25},
  node hidden/.style={node,blue!20!black,draw=myblue!30!black,fill=myblue!20},
  node convol/.style={node,orange!20!black,draw=myorange!30!black,fill=myorange!20},
  node out/.style={node,red!20!black,draw=myred!30!black,fill=myred!20},
  connect/.style={thick,mydarkblue}, %,line cap=round
  connect arrow/.style={-{Latex[length=4,width=3.5]},thick,mydarkblue,shorten <=0.5,shorten >=1},
  node 1/.style={node in}, % node styles, numbered for easy mapping with \nstyle
  node 2/.style={node hidden},
  node 3/.style={node out}
}
\def\nstyle{int(\lay<\Nnodlen?min(2,\lay):3)} % map layer number onto 1, 2, or 3

\graphicspath{{images/}}

\title{Resolving Spatio-Temporal Entanglement in Video Prediction via Multi-Modal Attention}

\author{
Shreyam Gupta$^{*1}$
\and
Pranjal Agrawal$^2$\And
Priyam Gupta$^3$\\
\affiliations
$^1$Indian Institute of Technology (BHU), Varanasi, India\\
$^2$University of Colorado, Boulder, USA\\
$^3$Erasmus+, Intelligent Field Robotic Systems (IFRoS), University of Girona, Spain\\
\emails
*shreyam.gupta.mec21@iitbhu.ac.in,
u1999097@campus.udg.edu
}
\begin{document}

\maketitle

\begin{abstract}
    The fast progress in computer vision has necessitated more advanced methods for temporal sequence modeling. This area is essential for the operation of autonomous systems, real-time surveillance, and predicting anomalies. As the demand for accurate video prediction increases, the limitations of traditional deterministic models, particularly their struggle to maintain long-term temporal coherence while providing high-frequency spatial detail, have become very clear. This report provides an exhaustive analysis of the Multi-Attention Unit Cell (MAUCell), a novel architectural framework that represents a significant leap forward in video frame prediction. By synergizing Generative Adversarial Networks (GANs) with a hierarchical ``STAR-GAN" processing strategy and a triad of specialized attention mechanisms (Temporal, Spatial, and Pixel-wise), the MAUCell addresses the persistent ``deep-in-time" dilemma that plagues Recurrent Neural Networks (RNNs). Our analysis shows that the MAUCell framework successfully establishes a new state-of-the-art benchmark, especially in its ability to produce realistic video sequences that closely resemble real-world footage while ensuring efficient inference for real-time deployment. Through rigorous evaluation on datasets: Moving MNIST, KTH Action, and CASIA-B, the framework shows superior performance metrics, especially in Learned Perceptual Image Patch Similarity (LPIPS) and Structural Similarity Index (SSIM). This success confirms its dual-pathway information transformation system. This report details the theoretical foundations, detailed structure and broader significance of MAUCell, presenting it as a valuable solution for video forecasting tasks that require high precision and limited resources.

\end{abstract}

\section{Introduction}
\colorlet{myred}{red!80!black}
\colorlet{myblue}{blue!80!black}
\colorlet{mygreen}{green!60!black}
\colorlet{myorange}{orange!70!red!60!black}
\colorlet{mydarkred}{red!30!black}
\colorlet{mydarkblue}{blue!40!black}
\colorlet{mydarkgreen}{green!30!black}

% STYLES
\tikzset{
  >=latex, % for default LaTeX arrow head
  node/.style={thick,circle,draw=myblue,minimum size=18,inner sep=0.4,outer sep=0.5},
  node in/.style={node,green!20!black,draw=mygreen!30!black,fill=mygreen!25},
  node hidden/.style={node,blue!20!black,draw=myblue!30!black,fill=myblue!20},
  node convol/.style={node,orange!20!black,draw=myorange!30!black,fill=myorange!20},
  node out/.style={node,red!20!black,draw=myred!30!black,fill=myred!20},
  connect/.style={thick,mydarkblue}, %,line cap=round
  connect arrow/.style={-{Latex[length=3.5,width=3]},thick,mydarkblue,shorten <=0.4,shorten >=0.8},
  node 1/.style={node in}, % node styles, numbered for easy mapping with \nstyle
  node 2/.style={node hidden},
  node 3/.style={node out},
  node rect/.style={
      thick,
      rectangle,
      draw=myred,
      fill=myred!20,
      minimum width=2.5cm,
      minimum height=1cm,
      align=center
    }
}
\def\nstyle{int(\lay<\Nnodlen?min(2,\lay):3)} % map layer number onto 1, 2, or 3

\begin{figure*}[t]

\centering
\resizebox{0.7\textwidth}{!}{%
% \hspace*{-2cm}
\begin{tikzpicture}[scale=1, x=1.5cm,y=0.9cm]
  % \large
  \message{^^JNeural network without arrows}
  \readlist\Nnod{5,4,3,2,3,4,5} % array of number of nodes per layer
  
  % TRAPEZIA
  \node[above,align=center,myorange!60!black] at (3,2) {Encoder};
  \node[above,align=center,myblue!60!black] at (5,2) {Decoder};
  \draw[fill=myorange,draw=none,fill opacity=0.00,rounded corners=2]
    (1.6,-2.7) --++ (0,5.4) --++ (2.8,-1.2) --++ (0,-3) -- cycle;
  \draw[draw=none,fill=myblue,fill opacity=0.00,rounded corners=2]
    (6.4,-2.7) --++ (0,5.4) --++ (-2.8,-1.2) --++ (0,-3) -- cycle;
  
  \message{^^J  Layer}
  \foreachitem \N \in \Nnod{ % loop over layers
    \def\lay{\Ncnt} % alias of index of current layer
    \pgfmathsetmacro\prev{int(\Ncnt-1)} % number of previous layer
    \message{\lay,}
    \foreach \i [evaluate={\y=\N/2-\i+0.5; \x=\lay; \n=\nstyle;}] in {1,...,\N}{ % loop over nodes
      
      % NODES
      \node[node \n,outer sep=0.4] (N\lay-\i) at (\x,\y) {};
      
      % CONNECTIONS
      \ifnum\lay>1 % connect to previous layer
        \foreach \j in {1,...,\Nnod[\prev]}{ % loop over nodes in previous layer
          \draw[connect,white,line width=1] (N\prev-\j) -- (N\lay-\i);
          \draw[connect] (N\prev-\j) -- (N\lay-\i);
          %\draw[connect] (N\prev-\j.0) -- (N\lay-\i.180); % connect to left
        }
      \fi % else: nothing to connect first layer
      
    }
  }
  
  % GAN STRUCTURE ADDITIONS
  % Replace latent vector and generator
  % \node[node in] (zin) [left=1.8 of N1-3] {$\vec z_\text{in}$};
  % \draw[connect arrow] (zin) -- node[above] {Generator $G(\cdot)$} (N1-3);

    % me dotted add
    \node[node in, below=of N\Nnodlen-5] (real) {};
    \node[below=0.5,align=center,mygreen!60!black] at (real) {Real Frames};
    \node[node rect, right=2.5 of N\Nnodlen-3] (D) at ($(N\Nnodlen-3)!0.5!(real)$) {Discriminator (FDU)};
    \node[node rect, above=2.5 of D] (loss) {Loss Calculation};
    % \node[node out] (D) [right=1.8 of N\Nnodlen-3] {$D(\cdot)$};

    \coordinate[right=2.5em of N\Nnodlen-3, circle, fill, inner sep=0.15em] (pt1);
    \coordinate[right=2.5em of real, circle, fill, inner sep=0.15em] (pt2);

    \draw[-, dashed] (pt1) edge[bend right] coordinate[circle, fill=orange, inner sep=1mm, pos=0.7] (pt3) (pt2);
  \draw[connect arrow] (N\Nnodlen-3) -- (pt1) (real) -- (pt2) (pt3) -- (D);
  \draw[connect arrow] (D) -- ++(0,2) node[midway, above, align=center] {Feedback} -- (loss);
  
  \draw[connect arrow] (loss) -- ++(0,2.15) -- ++(-2.3,0) node[midway, above] {Optimized Weights} -- ($(N\Nnodlen-1.north) + (0.1, 1.5)$);

%rough 
% ($(N1-1.north) + (0, 1.5)$) -- ($(N\Nnodlen-1.north) + (0, 1.5)$)
%   node[midway, above, font=\footnotesize\bfseries] {Generator (MAUCells)};
%   \draw[connect arrow] (xreal) -- ++(1.8,0) node[midway, above] {$p_\text{data}(x)$} -- ++(0,-2) -- (D);

% ($(zin.north) + (0, 0.5)$) -- ++(2, 0)
%   node[midway, above, font=\footnotesize\bfseries] {Generator}
%   -- ($(N\Nnodlen-3.north) - (2, -0.5)$);
  
  % Discriminator branch
  % \node[node out] (D) [right=1.8 of N\Nnodlen-3] {$D(\cdot)$};
  % \node[above=0.8 of D,align=center] {Discriminator};
  % \draw[connect arrow] (N\Nnodlen-3) -- node[above] {Generated Frames} (D);
  % \node[right=1.2 of D] (realfake) {Real?};
  % \draw[connect arrow] (D) -- (realfake);
  
  % Real data input
  % \node[node out] (xreal) [above=2.5 of zin] {$\vec x_\text{real}$};
  % \draw[connect arrow] (xreal) -- ++(1.8,0) node[midway, above] {$p_\text{data}(x)$} -- ++(0,-2) -- (D);

  % Labels
  \node[above=0.5,align=center,mygreen!60!black] at (N1-1.90) {Input Frames};
  \node[above=0.5,align=center,myred!60!black] at (N\Nnodlen-1.90) {Predicted Frames};
  \draw[thick, black]
  ($(N1-1.north) + (0, 1.5)$) -- ($(N\Nnodlen-1.north) + (0, 1.5)$)
  node[midway, above, font=\footnotesize\bfseries] {Generator (MAUCells)};
  
\end{tikzpicture}
}
\caption{Structure of the proposed GAN based system.}
\label{fig:gan}
\end{figure*}
The goal of video frame prediction is simple, given a sequence of past frames, a model needs to generate a sequence of future frames that are visually coherent and semantically plausible. However, this task requires an artificial system to learn the fundamental laws of physics, cause-and-effect relationships, and how object move without direct supervision. The field has gone through several distinct epochs, each defined by the key design approaches of the time.
In the nascent stages of computer vision, approaches relied heavily on optical flow estimation and shallow, pixel-based interpolation methods. These techniques, while computationally inexpensive, failed to capture complex, non-linear motions or handle occlusions effectively. The rise of deep learning brought in Recurrent Neural Networks (RNNs)~\cite{predrnn++,predrnn} and Long Short-Term Memory (LSTM)~\cite{e3d,srivastava2016unsupervisedlearningvideorepresentations,villegas2019highfidelityvideoprediction} networks. Models such as the ConvLSTM~\cite{NIPS2015_07563a3f,convlstm} expanded the fully connected LSTM to handle 2D spatial data. This change allowed for the extraction of spatial and temporal features at the same time. This was a significant moment. It allowed for the modeling of spatio-temporal dependencies in a unified framework.
However, standard RNN-based methods soon faced the ``deep-in-time" problem. As prediction horizons grew longer, gradients spread out over time would either fade away or become too large, causing to a loss of information about the initial state of the sequence. Additionally, models that used pixel-wise reconstruction losses, like Mean Squared Error (MSE), often fell into the trap of predicting the ``average" of all possible outcomes. In the context of stochastic video dynamics, this averaging showed up as blurriness i.e. a visual sign of the model's uncertainty. While the background might remain sharp, moving objects would lose focus and blur into a mix of probable averages.

\subsection{The Deterministic vs. Stochastic Divide}
A major divide in video prediction research exists between deterministic and stochastic modeling. Deterministic models predict a single future path based on past context. They work well for simple physics, like bouncing balls, but struggle in complex settings where many future scenarios are possible, such as a pedestrian nearing an intersection. Stochastic models, like Variational Autoencoders (VAEs) and probabilistic GANs, aim to model the distribution of possible futures.
The MAUCell framework has a unique role in this field. It is mainly structurally deterministic in how it generates inferences, but it also uses Generative Adversarial Networks (GANs)~\cite{savp,Kwon2019PredictingFF} to add stochastic realism to its output. By putting a Generator against a Discriminator, the model must create frames that fit the characteristics of real natural images, avoiding the blurry averages that mean squared error (MSE) tends to produce. This adversarial training acts as a proxy for stochastic sampling, promoting the creation of high-frequency details that deterministic methods usually smooth out.

\subsection{Limitations of Current State-of-the-Art}
Before MAUCell came into the picture, two main types of models dominated the field: advanced RNN variants such as PredRNN~\cite{predrnn,predrnnv2} and E3D-LSTM~\cite{e3d}, and newer Vision Transformers like SwinLSTM~\cite{swinLSTM}.
PredRNN and E3D-LSTM introduced complex memory systems (spatiotemporal memory flow) to help reduce the vanishing gradient problem. Still, they were heavy on computation and often had trouble separating motion features from appearance features. Their strictly sequential approach limited their ability to process in parallel, creating bottlenecks in training and inference speed.
SwinLSTM and Transformers: The use of the Swin Transformer for video prediction brought improved how well models understood global context. This was due to the self-attention mechanism's ability to model long-range dependencies. However, the quadratic complexity $O(N^2)$ of standard attention mechanisms makes them unsuitable for high-resolution video or real-time use on edge devices. While these models perform well in academic tests, their operational delays pose a significant challenge for real-world use. MAUCell addresses these specific gaps by proposing a ``lightweight" attention mechanism within an RNN framework. It aims to combine the long-range dependency capabilities of Transformers with the efficiency of convolutional recurrent networks.

\section{Architectural Framework: The STAR-GAN Paradigm}
The proposed solution is not just a new cell design. It also includes a clear processing strategy. The research presents a hierarchical processing architecture, called the STAR-GAN framework. This structure focuses on distributed feature computation across different levels. It separates immediate kinetic properties from lasting temporal paths.

\subsection{Hierarchical Video Prediction Generation}
The framework operates on a functional decoding mechanism. The generation of a future frame $\hat{v}_{t+1}$ is not a single-step inference but the result of a semantic reconstruction process initiated at the apex layer $N$ of a processing hierarchy.
The decoding function is defined as:
$$\hat{v}_{t+1} = \text{Dec}(S^{k=N}_t)$$
Here, $S^{k=N}_t$ represents the highest-level spatial state at time $t$. This state is the culmination of a bottom-up abstraction process. A loss calculation model contains integrated optimization metrics as illustrated by the figure~\ref{fig:gan} adapted from ~\cite{rnn,gan} .

\paragraph{The Spatial Foundation (Tier $k=0$):} At the base of the hierarchy, the system performs initial feature encoding. Raw video frames $v_t$ are transformed into structured spatial representations $S^{k=0}_t = \text{Enc}(v_t)$. This layer captures high-frequency details, such as edges, textures, and raw pixel intensities.
\paragraph{Distributed Computation (Tiers $k=1 \dots N$):} As information moves upward, each connected STAR-GAN module processes different parts of the sequence. Lower tiers focus on local motion and fine spatial details. Higher tiers capture global semantic shifts and long-term object paths.

A key innovation in this structure is the use of Feature Preservation Pathways. In traditional deep encoders, spatial resolution is often lost for semantic abstraction through downsampling. This loss is critical for video prediction, where the output needs to be a sharp image. The MAUCell architecture uses structural bridges to create direct links between encoder and decoder layers at the same resolutions. These pathways effectively avoid processing delays, ensuring that the fine-grained spatial information~\cite{akan2021slampstochasticlatentappearance} needed for sharp reconstruction is maintained throughout the hierarchical changes.

\subsection{Generator Module: Dual-Pathway Information Pipeline}
The main part of the STAR-GAN framework is the Generator. It does not simply pass a hidden state forward. Instead, it actively handles visual information using a Dual-Pathway Information Transformation Pipeline. This pipeline controls how data flows into two separate, dynamically managed repositories:
\paragraph{Appearance Information Store ($S$):} Dedicated to maintaining structural integrity and spatial detail.
\paragraph{Motion Information Store ($T$):} Dedicated to tracking temporal dynamics and kinetic flows.
Unlike standard RNNs where ``memory" is a passive vector, these stores function as resource allocation systems where data undergoes ``knowledge consolidation." At each processing tier $R^1$, the system performs a Contextual Relevance Calculation. It combines the current appearance representation with a sliding window of historical contexts:
Appearance Context: $S_{t-\tau+1 : t-1}$\\
Motion Context: $T_{t-\tau : t-1}$
This multi-modal integration allows the model to derive attention weights that consider both the current spatial configuration and the extended temporal history ($\tau$ frames)~\cite{stau}. The fusion operation produces updated matrices, a refined appearance state $S^1_t$ and a motion dynamics encoding $T^1_t$. Importantly, the motion encoding $T^1_t$ is not only used for the current step but is also added to the motion store to prepare the system for future prediction cycles, such as forecasting $F_{t+2}$). This creates a strong temporal continuity.

\begin{figure*}[t]
  \centering
  \includegraphics[width=\textwidth]{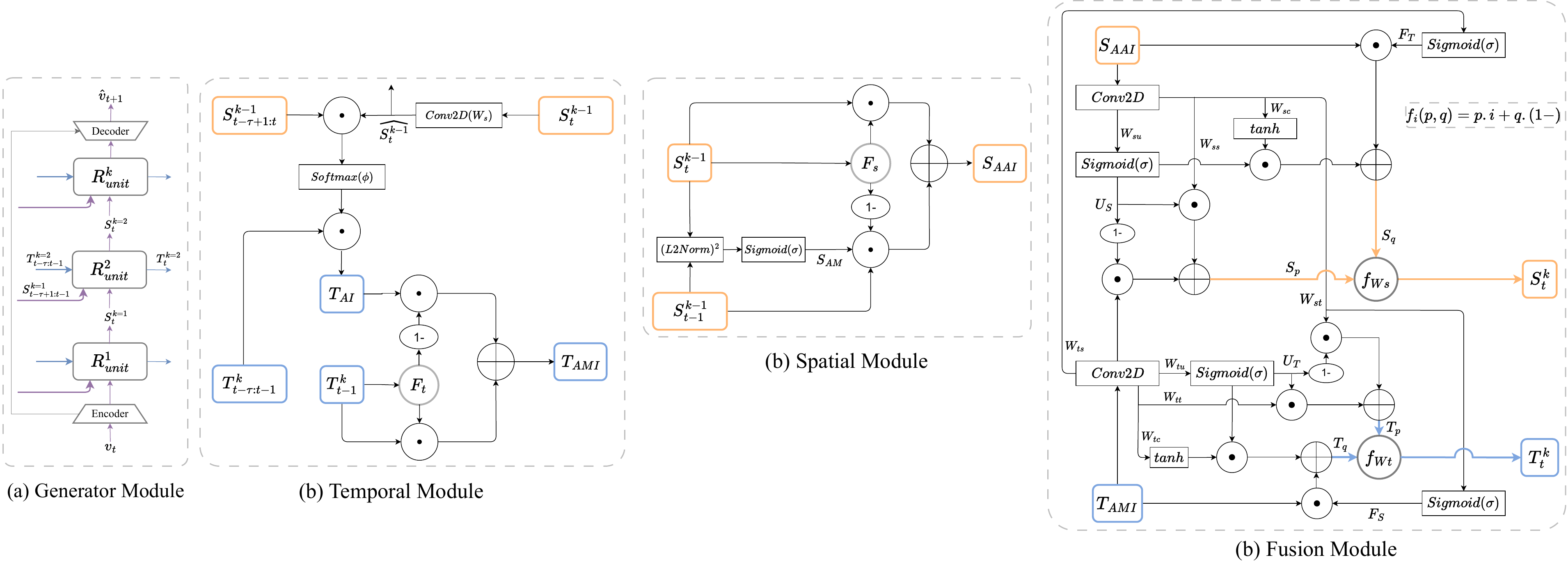}
  \caption{Multi Attentive Unit Cell (MAUCell) abstract recurrent unit structure, (a) Generator module with multiple predictive units. (b) Temporal attention module (c) Spatial attention module (d) Fusion module for combining spatial and the temporal information to produce final predictions.}
  \label{fig:memory}
\end{figure*}

\section{Multi-Attention Unit (MAU) Cell}
The MAUCell is the fundamental computational unit that replaces the standard LSTM cell within the RNN backbone. It solves the ``spatio-temporal entanglement" problem by treating spatial and temporal features as separate but interacting modalities. The cell combines three specialized attention mechanisms: Temporal Attention (TAM), Spatial Attention (SAM), and Pixel-wise Attention. These are fused through a Spatiotemporal-Aware Unit (STAM).

\begin{algorithm}[h]
\caption{Multi-Attention Unit Cell (MAUCell)}
\label{alg:maucell}
\begin{algorithmic}[1]
\REQUIRE Video frames \(T_t, S_t\), Attention weights \(t\_att, s\_att, s\_pixel\_att\)
\ENSURE Updated attention outputs \(\text{out\_T}, \text{out\_S}\)
\STATE Initialize model parameters \(\alpha_s, \alpha_t\) and hidden dimensions \(\Theta\)
\STATE Compute temporal and spatial attention projections:
\STATE \(s\_next = \text{Conv}(S_t)\), \(t\_next = \text{Conv}(T_t)\)
\STATE Aggregate temporal trends: \(T\_trend = \sum_{i=1}^\tau \text{Softmax} \left( \frac{s\_att[i] \cdot s\_next}{\sqrt{d}} \right) \cdot t\_att[i]\)
\STATE Compute gating mechanisms:
\STATE \(t\_att\_gate = \text{Sigmoid}(t\_next), \, s\_att\_gate = \text{Sigmoid}(s\_next)\)
\STATE Fuse features:
\STATE \(T\_fusion = T_t \cdot t\_att\_gate + (1 - t\_att\_gate) \cdot T\_trend\)
\STATE \(S\_fusion = S_t \cdot \text{Sigmoid}(s\_pixel\_att)\)
\STATE Split and apply activations:
\STATE \(t_i, t_r, t_t, t_s = \text{Split}(\text{Conv}(T\_fusion))\)
\STATE \(s_i, s_r, s_t, s_s = \text{Split}(\text{Conv}(S\_fusion))\)
\STATE Compute updated features:
\STATE \(T\_new\_1 = T_r \cdot T_i + S_t \cdot T\_fusion\)
\STATE \(S\_new\_1 = S_r \cdot S_i + T_s \cdot S\_fusion\)
\STATE Apply weighted fusion:
\STATE \(\text{out\_T} = \alpha_t \cdot T\_new\_1 + (1 - \alpha_t) \cdot T\_new\_2\)
\STATE \(\text{out\_S} = \alpha_s \cdot S\_new\_1 + (1 - \alpha_s) \cdot S\_new\_2\)
\RETURN \(\text{out\_T}, \text{out\_S}\)
\end{algorithmic}
\end{algorithm}

\subsection{Temporal Attention Module (TAM)}
The primary challenge in temporal modeling is maintaining context over long horizons without inflating parameter counts. Standard 3D-Convolutional networks attempt this by increasing kernel depth, which leads to massive computational overhead. The TAM solves this by functioning as a cross-domain knowledge transfer system.\\
\textbf{Mechanism of Action:}\\
The TAM uses spatial representations to guide temporal understanding. It works on the idea that not all historical frames matter the same for current predictions. For example, in a periodic motion like walking, frames from $t-5$ might be more relevant than frames from $t-1$.
\begin{itemize}
    \item Relevance Assessment: The module projects the current spatial state $S^{k-1}_t$ into a query vector $S'$. It then computes an attention score $P_j$ by taking the dot product of this query with historical spatial states $S^{k-1}_{t-j+1}$: $$P_j = S^{k-1}_{t-j+1} \cdot S'$$
    \item This operation effectively measures the semantic similarity between the current scene and past scenes.
    \item Softmax Normalization: These scores are normalized to produce attention weights $a_i$. This ensures the model focuses its computational resources on the most relevant historical moments: $$a_i = \frac{e^{P_i}}{\sum_{j=1}^\tau e^{P_j}}$$
    \item Trajectory-Aware Synthesis: The module gathers historical temporal states $T^k_{t-i}$ using these weights to form the Temporally-Integrated Attention Information ($T_{AI}$): $$T_{AI} = \sum_{i=1}^\tau a_i T^k_{t-i}$$
    \item Adaptive Gating: Finally, to balance this long-term historical trend ($T_{AI}$) with the immediate, momentary kinetic properties of the previous step ($T^k_{t-1}$), the system uses a Sequential Integration Controller ($F_t$). This sigmoid gate dynamically decides the fusion ratio: $$F_t = \sigma(W_t * T^k_{t-1})$$
    $$T_{AMI} = F_t \odot T^k_{(t-1)} + (1 - F_t) \odot T_{AI}$$
    \item The result, $T_{AMI}$ (Augmented Motion Information), is a strong representation that captures both immediate motion and long-range trajectory patterns. This approach effectively reduces the memory loss often seen in standard RNNs.
\end{itemize}

\subsection{Spatial Attention Module (SAM)}
While TAM addresses when events occur, SAM addresses where significant changes are happening. In video prediction, static backgrounds require less processing than moving foreground objects. The SAM operates as a Visual Saliency Discovery Mechanism.
\paragraph{Mechanism of Action:}
The SAM is predicated on appearance change detection. It assumes that the most information-rich regions of a frame are those that exhibit deviation from the previous time step.
\begin{itemize}
    \item Feature Evolution Tracking: The module calculates the ``feature evolution metric" by finding the squared $L_2$ norm difference between the current spatial state and the previous one. This creates a Spatial Attention Map ($S_{AM}$): $$S_{AM} = \sigma(||S^{k-1}_{t} - S^{k-1}_{t-1}||^2_2)$$
    \item This map effectively highlights kinetically active regions, edges of moving objects, deforming surfaces, or entering/exiting elements.
    \item Contextual Modulation: An adaptive spatial fusion gate $F_s$ is computed to modulate the integration of this dynamic map with the static spatial features: $$F_s = \sigma(W_s * S^{k-1}_t)$$
    \item Improved Visual Encoding: The final output, Improved Visual Representation ($S_{AAI}$), combines the raw spatial features with the motion-highlighted context: $$S_{AAI} = F_s \odot S_t^{k-1} + (1 - F_s) \odot S_{AM} \odot S^{k-1}_{t-1}$$
    
\end{itemize}
By focusing on areas with high feature deviation ($S_{AM}$), the SAM ensures that the model keeps fine-scale dynamic objects and structural edges that usually blur in standard convolutional down sampling.

\subsection{Pixel-wise Attention}
Operating at the most detailed level, the Pixel-wise Attention mechanism evaluates local importance by examining intensity changes on a pixel-by-pixel basis.
$$S_{\text{pixel\_att}} = ||S_t - S_{\text{prev}}||^2$$
 This component serves as a final refinement stage. It ensures that even small textural changes or minor object movements are not lost due to higher-level spatial attention masks. It is especially effective in maintaining perceptual consistency in complex textures such as water and foliage, where the overall structure may remain the same but local pixel intensities can vary greatly.

 \subsection{Multi-Stream Information Synthesis Framework (STAM)}
 The crowning feature of the MAUCell is the Spatiotemporal-Aware Unit (STAM), which synthesizes the outputs of the TAM ($T_{AMI}$) and SAM ($S_{AAI}$). Unlike conventional ``sequential fusion" approaches that merge motion and appearance in a linear step, the MAUCell employs a Parallel Information Integration System utilizing Quaternary Transformation Pathways.
 \begin{algorithm}[!t]
\caption{MAU-Based Sequential Video Prediction Framework}
\label{alg:mau_framework}
\begin{algorithmic}[1]
\REQUIRE Input frames $X = \{x_1, x_2, \dots, x_T\}$, Ground-truth mask $\text{Mask\_true}$, Model parameters $\Theta$ (encoders, decoders, MAUCells)
\ENSURE Predicted frames $\hat{X} = \{\hat{x}_{T+1}, \dots, \hat{x}_{T+H}\}$
\STATE Initialize temporal and spatial features $T_t, T_{\text{pre}}, S_{\text{pre}}, S_{\text{prev}}$ to zeros
\FOR{$t = 1$ \TO $T+H$}
    \IF{$t \leq T$}
        \STATE Set input frame: $x_t \gets X[t]$
    \ELSE
        \STATE Perform scheduled sampling:
        \[
        x_t \gets \text{Mask\_true}[t-T] \cdot x_t + (1 - \text{Mask\_true}[t-T]) \cdot \hat{x}_t
        \]
    \ENDIF
    \STATE Encode input frame through encoder layers: $S_t \gets \text{Encoders}(x_t)$
    \STATE Compute pixel-level attention:
    \[
    S_{\text{pixel\_att}} \gets ||S_t - S_{\text{prev}}||^2
    \]
    \STATE Update spatial features: $S_{\text{prev}} \gets S_t$
    \FOR{each MAUCell layer $i$}
        \STATE Gather temporal and spatial attention:
        \[
        t\_att \gets T_{\text{pre}}[i], \, s\_att \gets S_{\text{pre}}[i]
        \]
        \STATE Update features using MAUCell:
        \[
        T_t, S_t \gets \text{MAUCell}[i](T_t, S_t, t\_att, s\_att, S_{\text{pixel\_att}})
        \]
        \STATE Append updated features: $T_{\text{pre}}[i] \gets T_t, \, S_{\text{pre}}[i] \gets S_t$
    \ENDFOR
    \STATE Decode spatial features to predict frame:
    \[
    \hat{x}_t \gets \text{Decoders}(S_t)
    \]
    \STATE Store predicted frame: $\hat{X}[t] \gets \hat{x}_t$
\ENDFOR
\STATE Return all predicted frames: $\hat{X}$
\end{algorithmic}
\end{algorithm}

 \paragraph{Quaternary Pathways: }
 The enriched features ($T_{AMI}, S_{AAI}$) are processed through four distinct parallel convolution streams. This design philosophy acknowledges that spatial and temporal features interact in complex, non-linear ways.
 \begin{itemize}
     \item Intra-Modal Streams: $T'_{tt}$ (Temporal-to-Temporal) and $S'_{ss}$ (Spatial-to-Spatial).
     \item Cross-Modal Streams: $S'_{st}$ (Spatial-to-Temporal) and $T'_{ts}$ (Temporal-to-Spatial).
 \end{itemize}

 \paragraph{Dual-Pathway Integration Strategy:}
The system uses two strategies to combine these streams:
\begin{itemize}
    \item Primary Integration Strategy ($T_p, S_p$): This strategy focuses on immediate, active merging with update gates ($U_T, U_S$). It synthesizes cross-domain features to produce spatially improved temporal encodings and temporally informed spatial maps.
    $$U_T = \sigma(W_{tu} * T_{AMI}), \quad U_S = \sigma(W_{su} * S_{AAI})$$
    $$T_p = U_T \odot T'_{tt} + (1 - U_T) \odot S'_{st}$$
    $$S_p = U_S \odot S'_{ss} + (1 - U_S) \odot T'_{ts}$$
    \item Complementary Integration Strategy ($T_q, S_q$): This strategy focuses on memory retention and pixel consistency. It uses tanh activations and filter gates ($F_S, F_T$) to model stable, long-term dependencies that need be preserved despite immediate changes.
    $$T_q = F_S \odot T_{AMI} + \tanh(T'_{tc}) \odot U_T$$
    $$S_q = F_T \odot S_{AAI} + \tanh(S'_{sc}) \odot U_S$$
\end{itemize}

 \paragraph{Parametric Blending:}
 The final representations for layer $k$ at time $t$ are generated via a learned interpolation mechanism. Learnable parameters $\alpha_t$ and $\alpha_s$ dynamically weigh the roles of the Primary (active) and Complementary (stable) strategies:
$$T^k_t = \alpha_t \odot T_p + (1 - \alpha_t) \odot T_q$$
$$S^k_t = \alpha_s \odot S_p + (1 - \alpha_s) \odot S_q$$
This architecture allows the network to effectively ``choose" between relying on immediate motion cues or established memory states. This provides a flexibility that rigid LSTM architectures do not have.

\section{Adversarial Training and Discriminator Design}
A key feature of the MAUCell framework is its use of Generative Adversarial Networks (GANs). Standard video prediction models trained on reconstruction losses (L1/L2) face the ``regression to the mean" problem, leading to blurry predictions. The MAUCell combats this by introducing a Frame Discriminator Unit (FDU).

\subsection{The Frame Discriminator Unit (FDU)}
The FDU is not a typical image discriminator; it evaluates spatio-temporal consistency. It must tell apart a ``real" video sequence and a ``generated" one.
\begin{algorithm}[!t]
\caption{Frame Discriminator Unit (FDU)}
\label{alg:fdu}
\begin{algorithmic}[1]
\REQUIRE Input frames $X = \{x_1, x_2, \dots, x_T\}$ (real or generated), Model parameters $\Theta$ (encoders, MAUCells)
\ENSURE Discrimination results $Y = \{y_1, y_2, \dots, y_T\}$ (real or fake classification for each frame)
\STATE Initialize temporal and spatial features $T_t, T_{\text{pre}}, S_{\text{pre}}, S_{\text{prev}}$ as zero tensors
\FOR{$t = 1$ \TO $T$}
    \STATE Extract spatial features $S_t$ using encoder layers:
    \[
    S_t = \text{Encoders}(x_t)
    \]
    \STATE Compute pixel-level attention for frame $S_t$:
    \[
    S_{\text{pixel\_att}} = ||S_t - S_{\text{prev}}||^2
    \]
    \STATE Update spatial features: $S_{\text{prev}} \gets S_t$
    \FOR{each MAUCell layer $i$}
        \STATE Gather temporal and spatial attention:
        \[
        t\_att \gets T_{\text{pre}}[i], \, s\_att \gets S_{\text{pre}}[i]
        \]
        \STATE Update features using MAUCell:
        \[
        T_t, S_t = \text{MAUCell}[i](T_t, S_t, t\_att, s\_att, S_{\text{pixel\_att}})
        \]
        \STATE Append updated features:
        \[
        T_{\text{pre}}[i] \gets T_t, \, S_{\text{pre}}[i] \gets S_t
        \]
    \ENDFOR
    \STATE Compute classification for frame $x_t$:
    \[
    y_t = \text{Decoder}(S_t, T_t)
    \]
    \STATE Store classification result: $Y[t] \gets y_t$
\ENDFOR
\RETURN Discrimination results $Y$
\end{algorithmic}
\end{algorithm}

\paragraph{Structure:} The discriminator matches the generator's attention-based design. It uses multi-scale extraction to assess both basic and advanced frame features.Function: It provides a supervisory signal that penalizes the generator not just for pixel errors, but for producing ``fake-looking" motion artifacts or structurally incoherent objects. This forces the generator to lie on the manifold of natural videos.

\subsection{Hybrid Loss Function}
The training objective combines three different loss components by weights to balance structural fidelity and perceptual realism:
$$L_{gen} = \alpha \cdot L_{MSE} + \beta \cdot L_{L1} + \gamma \cdot L_{adv}$$
MSE Loss ($L_{MSE}$) optimizes for structural accuracy and high PSNR. It penalizes large deviations in pixel intensity.
L1 Loss ($L_{L1}$) minimizes absolute deviations, improving textural content and reducing the ``blur" linked to MSE.
Adversarial Loss ($L_{adv}$) comes from the Binary Cross-Entropy (BCE) of the discriminator. It pushes the generator to create frames that look like real data, enhancing perceptual quality (LPIPS).
\paragraph{Discriminator Loss: }$$L_{disc} = \text{BCE}(D_{real}, 1) + \text{BCE}(D_{fake}, 0)$$
This adversarial setup ensures that the MAUCell not only memorize patterns but also learns to create believable continuations of visual reality.

\section{Experimental Validation}
The MAUCell framework underwent thorough evaluation against several current baselines using three benchmark datasets. The evaluation method emphasized a full view of performance, combining traditional metrics with indicators of perceptual quality.

\subsection{Benchmark Datasets} The chosen datasets aimed to address specific challenges in video prediction:

\paragraph{Moving MNIST~\protect\cite{deng2012mnist,lee2019mutualsuppressionnetworkvideo}:}Synthetic.
Two handwritten digits moving with random trajectories in a $64 \times 64$ frame.
Challenge: Occlusion and Entanglement. The digits often cross paths, requiring the model to track overlapping objects and separate their features. The ``random" trajectories test the model's ability to learn non-linear physics without semantic context.
Scale: 10,000 sequences (7,000 training, 3,000 testing).

\paragraph{KTH Action~\protect\cite{kth}:}Real-world.
5 subjects perform six human actions (walking, jogging, running, boxing, waving, clapping) in various environments.\\
Challenge: Articulated Motion and Background Stability. The model must predict complex limb movements while keeping the background stable. This tests the model’s ability to differentiate between the dynamics of the foreground and the stability of the background.\\
Scale: 5,686 training sequences, 2,437 test sequences.

\paragraph{CASIA-B (Preprocessed)~\protect\cite{casiab_public,casiabp}:}Real-world.
Large-scale gait database with subjects filmed from multiple viewing angles.
Challenge: Viewpoint Variation and Fine-Grained Detail. This is the most rigorous test of the Spatial Attention Module. The model needs to generalize motion patterns from side views, frontal views, and oblique angles, while also dealing with occlusions from carrying bags or wearing coats.
Scale: Extensive multi-view sequences.

\subsection{Comparative Analysis}
The MAUCell was tested against established RNN models (PredRNN, PredRNN-V2, E3D-LSTM), CNN models (SimVP~\cite{simvp}, CrevNet~\cite{crevnet}), and Transformer models (SwinLSTM).

\subsubsection{Quantitative Performance on Moving MNIST}
\begin{table*}[t]
    \centering
    \begin{tabularx}{\linewidth}{p{6cm}XXXXX}
        \toprule
        Model & PSNR$\uparrow$ & SSIM$\uparrow$ & LPIPS$\downarrow$ & MSE$\downarrow$ & Computation Time (s)\\ \midrule
        \textbf{PredRNN}~~\cite{predrnn} & 18.1 & 0.88 & 6.9 & 75.1 & 59.00 \\
        \textbf{PredRNN-V2}~~\cite{predrnnv2} & 17.9 & 0.879 & 6.8 & 74.2 & 51.50 \\
        \textbf{Eidetic 3D LSTM}~~\cite{e3d}& 17.5 & 0.749 & 17 & 90.1 & 31.00 \\
        \textbf{MMVP}~~\cite{mmvp} & 15.3 & 0.802 & 19.3 & 89.1 & 30.10 \\
        \textbf{CrevNet}~~\cite{crevnet} & 17 & 0.76 & 20 & 97.8 & 23.80 \\
        \textbf{LMCNet}~~\cite{lmcnet} & 17.5 & 0.76 & 17.1 & 92 & 23.00 \\
        \textbf{PhyDnet}~~\cite{phydnet} & 19.3 & 0.837 & 14.8 & 62.5 & 20.60 \\
        \textbf{SVG}~~\cite{svg} & 17.7 & 0.872 & 8.2 & 81 & 19.90 \\
        \textbf{MAU}~~\cite{mau} & 19.6 & 0.91 & 5.98 & 48 & 19.80 \\
        \textbf{SAVP}~~\cite{savp} & 18.2 & 0.909 & 6.13 & 28.61 & 18.90 \\
        \textbf{SwinLSTM}~~\cite{swinLSTM} & \textbf{43.1} & \textbf{0.971} & \textbf{2.8} & \textbf{14.7} & 18.10\\
        \textbf{STAU}~~\cite{stau} & 18.5 & 0.885 & 5.3 & 25.41 & 16.00 \\ \hline
        Proposed Model & 22.5 & 0.935 & 4.8 & 43.5 & \textbf{14.10} \\ \bottomrule
    \end{tabularx}
    \caption{Quantitative comparison of video prediction models on the Moving MNIST dataset.}
    \label{tab:mnist}
\end{table*}

The results on Moving MNIST highlight the MAUCell's efficiency and perceptual superiority.
\paragraph{Efficieny:}
MAUCell achieves an inference time of 14.1 seconds, the fastest among all compared models. It is ~4.2x faster than PredRNN and ~22\% faster than the Transformer-based SwinLSTM. This validates the ``lightweight" design philosophy of the MAUCell, achieving complex attention without the heavy computational cost of global self-attention mechanisms.
\paragraph{Perceptual Quality:}     The LPIPS score of 4.8 is excellent for an RNN-based model. It outperforms STAU (5.3) and PredRNN (6.9). This shows that the GAN component effectively generates sharp, realistic frames. It avoids the blurriness that raises MSE but lowers visual quality.

\subsubsection{Performance on KTH Action Dataset}

On real-world human motion data, the MAUCell demonstrates clear dominance.
\paragraph{Insight} The MAUCell significantly outperforms SwinLSTM on KTH (PSNR 38.5 vs 34.8). This suggests that for biological motion, the Temporal Attention Module (TAM), which explicitly tracks motion trends over time windows ($\tau$). It performs better than the generic global attention found in Transformers. The extremely low MSE of 0.63 shows high pixel-level precision. Meanwhile, the near-perfect SSIM of 0.991 confirms that structural details are well preserved.

\subsubsection{Performance on CASIA-B Dataset}
The CASIA-B results underscore the model's robustness to viewpoint changes.
\paragraph{Insight} This dataset reveals the critical contribution of the Spatial Attention Unit (SAU). SwinLSTM has difficulty in this area, achieving an SSIM of 0.521. This is probably due to standard attention mechanisms not being able to handle large viewpoint changes without noticeable spatial gating. The MAUCell, which includes a mechanism for discovering visual saliency, keeps structural coherence with an SSIM of 0.843. It does this by adaptively focusing on stable features of the gait cycle while ignoring background noise and perspective distortions.

\subsection{Ablation Studies} To isolate the impact of the GAN component, ablation studies were conducted on the Moving MNIST dataset.
\textbf{Analysis:}\\
L1 Only: The model produces structurally sound but blurry images (high LPIPS of 16.5). It learns the ``average" motion but lacks definition.\\
GAN Only: The model collapses. Without the guidance of reconstruction loss (L1/MSE) to ground the predictions in the ground truth, the GAN hallucinates realistic-looking but incorrect frames (low PSNR, low SSIM).\\
Full Synergy: The combination is greater than the sum of its parts. MSE/L1 provide the structural scaffold, while the GAN ``polishes" the texture and edges. The drop in LPIPS from 16.5 (L1) to 4.7 (Full) shows the huge improvement in perceptual realism due to the adversarial training.

\section{Broader Implications and Future Outlook}
\subsection{Real-Time Applications and Edge AI}
The most important practical finding is the MAUCell's inference speed of 14.1 seconds. In autonomous driving or drone navigation, latency is a critical safety issue. Transformer-based models, while powerful, often require server-grade GPUs to run efficiently. The MAUCell's architecture leveraging the efficiency of RNNs with the power of attention, makes it a viable candidate for Edge AI deployment. It offers a ``sweet spot" between the lightweight speed of simple CNNs and the global reasoning of heavy Transformers.

\subsection{The ``STAR-GAN" Paradigm Shift}
The hierarchical ``STAR-GAN" structure represents shift toward modular, clear deep learning. By clearly separating ``appearance" and ``motion" streams and processing them through specialized levels, the model becomes easier to understand. We can inspect the ``Motion Information Store" to understand what dynamics the model is tracking, or visualize the ``Spatial Attention Maps" to see where the model is looking. This contrasts with the ``black box" nature of massive end-to-end Transformers.

\subsection{Limitations and Future Work}
While the MAUCell performs well in deterministic benchmarks, its current design mainly handles stochasticity through the GAN's distributional matching. It does not explicitly model a probability distribution over future states like VAE.\\
Stochastic Extension: Future versions could include latent variables (VAE-GAN)~\cite{stau,predrnn,svg,savp,babaeizadeh2018stochasticvariationalvideoprediction}. This would let the model to generate multiple diverse and realistic futures for a single input sequence. This capability is essential for long-horizon predictions in unpredictable environments, such as traffic intersections.\\
Multimodal Integration: The framework's modular structure makes it perfect for growth into multimodal learning. Adding audio (like engine sounds) or depth data (such as LiDAR) to the ``Information Stores" could greatly improve prediction accuracy in complex, real-world situations.
\begin{table*}[t]
    \centering
    \begin{tabularx}{\linewidth}{XXXXXX}
        \toprule
        Model & PSNR$\uparrow$ & SSIM$\uparrow$ & LPIPS$\downarrow$ & MAE$\downarrow$ & MSE$\downarrow$ \\ \midrule
        MAU & 35.1 & 0.98 & 3.57 & 40.3 & 0.93 \\
        PredRNN-V2 & 26.85 & 0.946 & 5.97 & 77.7 & 8.02 \\
        SwinLSTM & 34.8 & 0.897 & 3.22 & 20.21 & 6.02 \\ \hline
        Proposed Model & \textbf{38.5} & \textbf{0.991} & \textbf{1.97} & \textbf{33.02} & \textbf{0.63} \\ \bottomrule
    \end{tabularx}
    \caption{Quantitative comparison of video prediction models on the KTH Action dataset.}
    \label{tab:kth}
\end{table*}
\begin{table*}[t]
    \centering
    \begin{tabularx}{\linewidth}{XXXXXX}
        \toprule
        Model & PSNR$\uparrow$ & SSIM$\uparrow$ & LPIPS$\downarrow$ & MAE$\downarrow$ & MSE$\downarrow$ \\ \midrule
        MAU & 23.04 & 0.759 & 23.81 & 171 & 19.5 \\
        SwinLSTM & 20.02 & 0.521 & 38.20 & 291 & 46.6 \\ \hline
        Proposed Model & \textbf{25.24} & \textbf{0.843} & \textbf{21.11} & \textbf{162} & \textbf{13.5} \\ \bottomrule
    \end{tabularx}
    \caption{Quantitative comparison of video prediction models on the CASIA-B dataset.}
    \label{tab:casiab}
\end{table*}
\begin{table*}[b]
    \centering
    \begin{tabularx}{\linewidth}{XXXXXX}
        \toprule
        Loss & PSNR$\uparrow$ & SSIM$\uparrow$ & LPIPS$\downarrow$ & MAE$\downarrow$ & MSE$\downarrow$ \\ \midrule
        Reconstruction~(L1) & 15.7 & 0.75 & 16.5 & 165 & 91 \\
        GAN & 13.3 & 0.62 & 13.1 & 243 & 185 \\
        GAN+L1  & 11.4 & 0.71 & 23.2 & 311 & 238 \\
        GAN+L1+MSE & \textbf{19.8} & \textbf{0.88} & \textbf{4.7} & \textbf{72} & \textbf{38} \\ \hline
    \end{tabularx}
    \caption{Ablation studies of the GAN component on the MNIST data.}
    \label{tab:ablation}
\end{table*}

% \begin{figure*}[t]
%     \centering
%     \includegraphics[width=\linewidth]{images/MNIST_fig.png}
%     \caption{Visual comparison of generated frames of the Moving MNIST dataset.}
%     \label{fig:sample}
% \end{figure*}

% \begin{figure*}[t]
%     \centering
%     \includegraphics[width=\linewidth]{images/fig_KTH.png}
%     \caption{Visual comparison of generated frames of the KTH action dataset.}
%     \label{fig:sample}
% \end{figure*}

% \begin{figure*}[b]
%     \centering
%     \includegraphics[width=\linewidth]{images/fig_casia.png}
%     \caption{Visual comparison of generated frames of the CASIA-B dataset.}
%     \label{fig:sample}
% \end{figure*}
\begin{figure*}[p] % 'p' encourages a dedicated page for these floats
    \centering
    
    % First Image
    \begin{subfigure}{\linewidth}
        \centering
        \includegraphics[width=\linewidth]{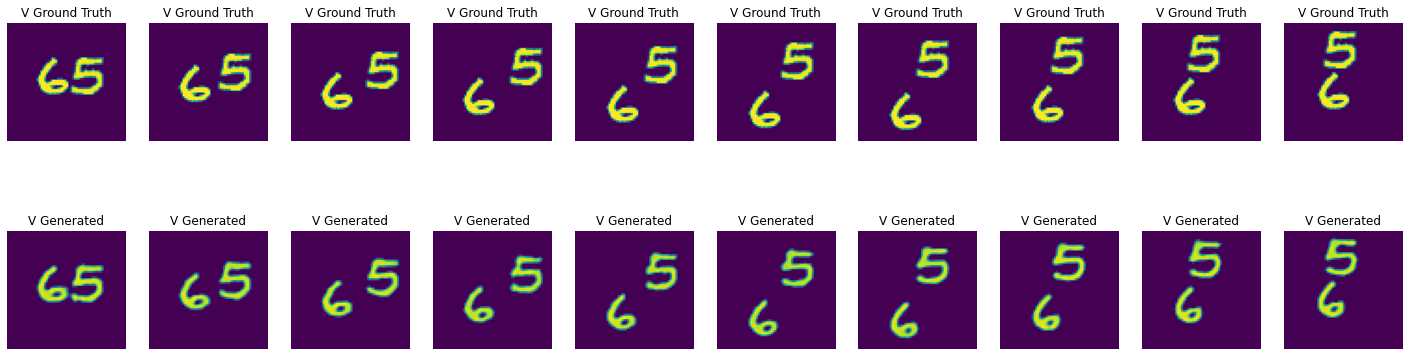}
        \caption{Visual comparison on the Moving MNIST dataset.}
        \label{fig:mnist}
    \end{subfigure}
    
    \vspace{1em} % Add some vertical space between images

    % Second Image
    \begin{subfigure}{\linewidth}
        \centering
        \includegraphics[width=\linewidth]{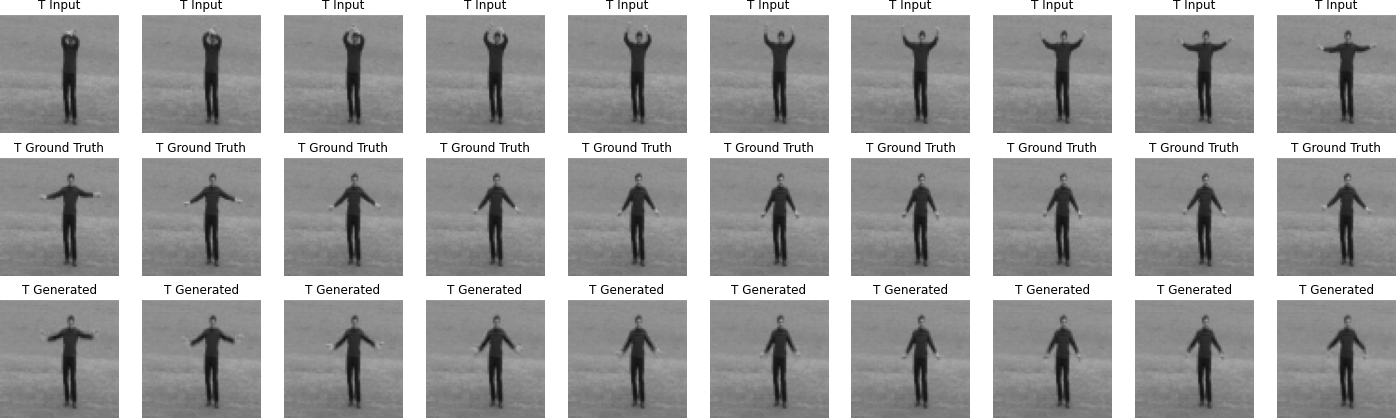}
        \caption{Visual comparison on the KTH action dataset.}
        \label{fig:kth}
    \end{subfigure}
    
    \vspace{1em}

    % Third Image
    \begin{subfigure}{\linewidth}
        \centering
        \includegraphics[width=\linewidth]{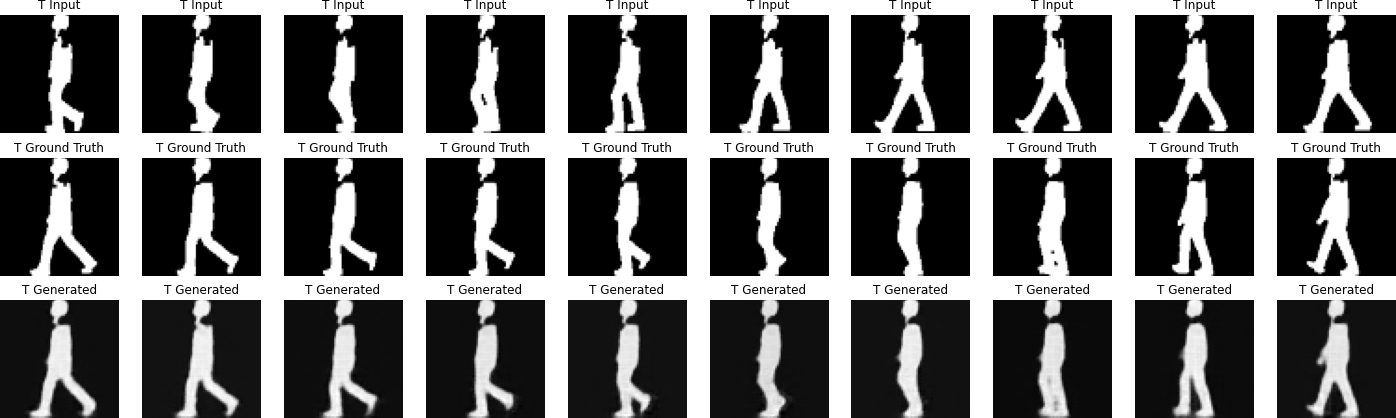}
        \caption{Visual comparison on the CASIA-B dataset.}
        \label{fig:casia}
    \end{subfigure}

    \caption{Qualitative results across different datasets. Our model maintains high-frequency details and temporal coherence.}
    \label{fig:all_results}
\end{figure*}

\section{Conclusion}
The MAUCell framework offers a solid solution to the complex challenges of video frame prediction. By combining the generative strength of GANs with the precise structure of a hierarchical Multi-Attention architecture, it successfully navigates the trade-offs between temporal coherence, spatial fidelity, and computational efficiency.\\
The empirical evidence is compelling:\\
Superior Dynamics Modeling: The Temporal Attention Module (TAM) effectively captures long-range dependencies without the computational cost of Transformers.\\
Visual Saliency: The Spatial Attention Unit (SAM) ensures that fine details and object shapes remain intact even in complex motion and viewpoint changes.
Operational Viability: With state-of-the-art inference speeds and excellent perceptual quality metrics, the MAUCell is not just a theoretical novelty but a practical tool ready for use in next-generation intelligent systems.\\

As the field moves toward more complex and unstructured environments, the MAUCell's ``STAR-GAN" design philosophy prioritizing modularity, efficiency, and clear spatio-temporal separation, will likely become a key template for future developments in temporal sequence modeling.

% \input{train_frame}
% Insert Table

%% The file named.bst is a bibliography style file for BibTeX 0.99c

\bibliographystyle{named}
\bibliography{references}

\clearpage
\appendix
\section{Appendix}
\subsection{Hardware Specification}
% \begin{table*}[!htbp]
%     \centering
%     \begin{tabularx}{\textwidth}{|p{3.5cm}|X|}
%         \hline
%         System Processor & AMD Ryzen Threadripper 3960X 24-Core Processor @ 3.8Ghz \\ \hline
%         Graphics & GeForce RTX 3090 with 24 GB RAM \\ \hline
%         System Memory & 64 GB Ram \\ \hline
%     \end{tabularx}
%     \caption{System Hardware}
%     \label{tab:hard_spec}
% \end{table*}

\centering
\begin{tabularx}{\textwidth}{|p{3.5cm}|X|}
    \hline
    System Processor & AMD Ryzen Threadripper 3960X 24-Core Processor @ 3.8Ghz \\ \hline
    Graphics & GeForce RTX 3090 with 24 GB RAM \\ \hline
    System Memory & 64 GB Ram \\ \hline
\end{tabularx}

\captionof{table}{System Hardware} % Requires \usepackage{caption}
\label{tab:hard_spec}

\end{document}